\documentclass[10pt,twocolumn,letterpaper]{article}
\usepackage[pagenumbers]{cvpr} % To force page numbers, e.g. for an arXiv version

\usepackage{comment}
\usepackage{multirow}
\usepackage{tabularx}
\usepackage{booktabs}
\usepackage{array}
 \usepackage{arydshln}
\usepackage{svg}        
\usepackage{verbatim}
\usepackage{pifont}
\usepackage{amsmath}
\usepackage{siunitx}
\usepackage{adjustbox}
\definecolor{cvprblue}{rgb}{0.21,0.49,0.74}
\usepackage[pagebackref,breaklinks,colorlinks,allcolors=cvprblue]{hyperref}

\def\paperID{11} % *** Enter the Paper ID here
\def\confName{CVPR}
\def\confYear{2026}

\title{LiveStre4m: Feed-Forward Live Streaming of \\ Novel Views from Unposed Multi-View Video}

\author{
Pedro Quesado \quad
Erkut Akdag \quad
 Yasaman Kashefbahrami \quad
 Willem Menu \quad
Egor Bondarev \\
AIMSGroup, Department of Electrical Engineering, Eindhoven University of Technology \\
{\tt\small \{p.quesado.dos.santos, e.akdag, y.kashefbahrami, w.j.menu, e.bondarev\}@tue.nl}
}

\begin{document}
\maketitle
\begin{abstract}
Live-streaming Novel View Synthesis (NVS) from unposed multi-view video remains an open challenge in a wide range of applications. Existing methods for dynamic scene representation typically require ground-truth camera parameters and involve lengthy optimizations ($\approx 2.67$s), which makes them unsuitable for live streaming scenarios. To address this issue, we propose a novel viewpoint video live-streaming method (LiveStre4m), a feed-forward model for real-time NVS from unposed sparse multi-view inputs. LiveStre4m introduces a multi-view vision transformer for keyframe 3D scene reconstruction coupled with a diffusion-transformer interpolation module that ensures temporal consistency and stable streaming. In addition, a Camera Pose Predictor module is proposed to efficiently estimate both poses and intrinsics directly from RGB images, removing the reliance on known camera calibration information. Our approach enables temporally consistent novel-view video streaming in real-time using as few as two synchronized unposed input streams. LiveStre4m attains an average reconstruction time of $ 0.07$s per-frame at $ 1024 \times 768$ resolution, outperforming the optimization-based dynamic scene representation methods by orders of magnitude in runtime. These results demonstrate that LiveStre4m makes real-time NVS streaming feasible in practical settings, marking a substantial step toward deployable live novel-view synthesis systems. \textit{Code available at:} \url{https://github.com/pedro-quesado/LiveStre4m}

%as egor has:
%Live-streaming Novel View Synthesis (NVS) from unposed multi-view video remains an open challenge in a wide range of applications. Existing methods for dynamic scene representation typically require ground-truth camera parameters and involve lengthy optimizations ($\approx 2.67$s), which makes them unsuitable for live streaming scenarios. To address this issue, we propose a novel viewpoint video live-streaming method (LiveStre4m), a feed-forward model for real-time NVS from unposed sparse multi-view inputs. LiveStre4m introduces a multi-view vision transformer for keyframe 3D scene reconstruction with a diffusion-transformer interpolation module, to ensure temporal consistency and stable streaming. In addition, a Camera Predictor module is proposed to efficiently estimate both poses and intrinsics directly from RGB images, removing the reliance on known camera calibration information. Our approach enables temporally consistent novel-view video streaming in real-time using as few as two synchronized unposed input streams. LiveStre4m attains an average reconstruction time of $ 0.08$s per-frame at $ 1152 \times 768$ resolution, outperforming the optimization-based dynamic scene representation methods by orders of magnitude in runtime. These results demonstrate that LiveStre4m makes real-time NVS streaming feasible in practical settings, marking a substantial step toward deployable live novel-view synthesis systems.

\end{abstract}

% \begin{abstract}
% Live-streaming Novel View Synthesis (NVS) from unposed multi-view video remains an unsolved task with a wide range of applications. Existing dynamic scene representation methods rely on long per-frame optimizations ($\approx 2.67$s) and require ground-truth camera information, which limits their applicability to real-world live streaming scenarios. We present LiveStre4m, a feed-forward framework for real-time NVS from unposed multi-view input. LiveStre4m integrates a multi-view vision transformer for key-frame 3D scene reconstruction with a diffusion transformer–based interpolation module to ensure temporal consistency and sufficient frame rate. Additionally, a Camera Predictor module estimates camera poses and intrinsics directly from RGB inputs eliminating the need for known camera parameters. The proposed approach enables temporally consistent novel-view video streaming in real-time from as few as two time-synchronized unposed input videos. It achieves real-time rendering, with an average reconstruction runtime of $ 0.08$s per-frame at a resolution of $ 640 \times 512$ with a small latency buffer of $\approx 1.06$s. Our experiments demonstrate that our method achieves photorealistic rendering while operating orders of magnitude faster than optimization-based baselines. Our method represents a significant step towards live-streaming novel view-point video.

% \end{abstract}    
\section{Introduction}
\label{sec:intro}
Synthesis and live-streaming of videos from novel viewpoints of a dynamic scene is of practical importance for applications in 3D scene understanding, sports broadcasting, and augmented reality. These applications condition the input to sparse unposed multi-view video streams. Despite substantial progress in Novel View Synthesis (NVS) for dynamic environments, most existing methods require ground-truth camera parameters and involve lengthy per-scene optimization, making them impractical for real-time synthesis and rendering.

%Live-streaming novel viewpoints of dynamic scenes from sparse unposed multi-view input is of practical relevance with applications in 3D scene understanding, sports broadcasting, and augmented reality. Despite substantial progress in Novel View Synthesis (NVS) for dynamic environments, most existing methods require ground-truth camera parameter and involve lengthy per-scene optimization, making them impractical for real-time rendering. 

Recent advances like 3D Gaussian Splatting (3DGS)~\cite{3dgs} enable efficient, photorealistic rendering of static scenes in real time. However, constructing the scene representation still requires several minutes of optimization. Inspired by 3DGS, extensions to dynamic scenes have been proposed~\cite{4dgs, 4dgauss, spacetime_gauss}, yet these methods rely on \textit{offline optimization} over the full multi-view video sequences. This optimization process prevents real-time deployment, making such methods unsuitable for live-streaming synthetic videos from novel viewpoints. More recent approaches adopt \textit{online per-frame optimization}~\cite{3dgstream, igs}, enabling streaming longer durations of novel videos. However, per-frame reconstruction remains slow, requiring several seconds per frame, thereby preventing real-time novel-video streaming. 

\begin{figure}[t]
  \centering
  \includegraphics[width=0.95\columnwidth]{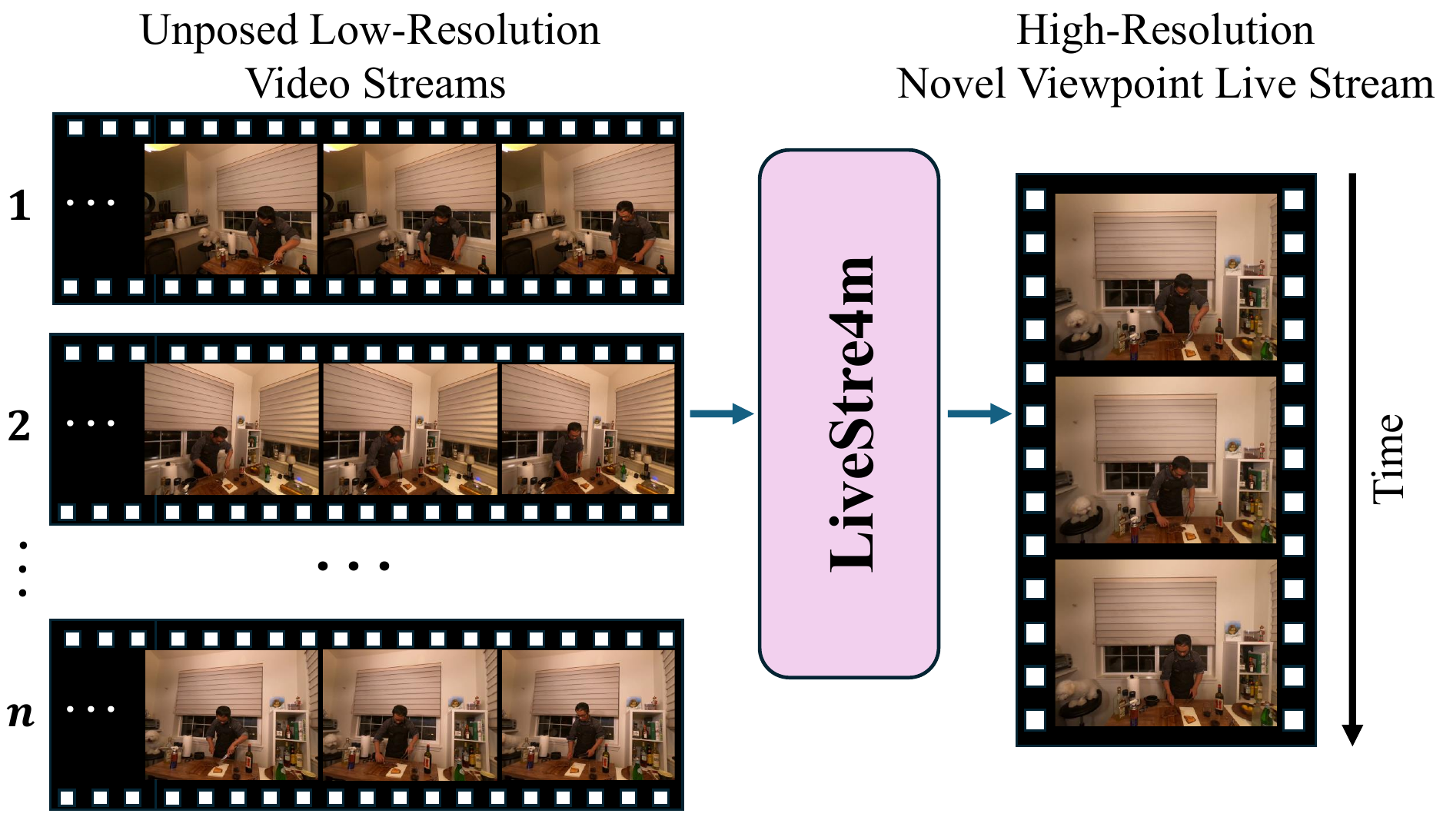}
  \caption{Illustration of the proposed LiveStre4m method, a feed-forward model for live-streaming novel viewpoint video from two or more low-resolution input streams.}
  \label{fig:livestre4m}
\end{figure}

Transformer-based architectures have recently reshaped the multi-view stereo and 3D reconstruction fields. For instance, DUSt3R~\cite{dust3r} and MASt3R~\cite{mast3r} leverage pre-trained Vision Transformers (ViTs)~\cite{ViT} to provide dense geometric correspondences and scene structure, reducing the reliance on conventional optimization-based pipelines. Building on these advancements, several feed-forward approaches for NVS of static scenes have emerged~\cite{splatt3r, FLARE, pixelsplat, quark}, achieving generalizable zero-shot photorealistic rendering without per-scene optimization. However, these methods generate each frame independently, offering limited temporal consistency and lacking requirements needed for streaming of novel viewpoint videos of dynamic scenes.

Another challenge for current NVS methods is the strong dependence on accurate ground-truth camera parameters. In real-world settings, reliable calibration data is unavailable and acquiring these parameters requires either specialized hardware (e.g., OptiTrack) or computationally expensive structure-from-motion pipelines, such as COLMAP~\cite{colmap1, colmap2}. Moreover, small calibration errors can lead to noticeable geometric distortion in generated novel views. Finally, diversity in camera parameter conventions across datasets hinders robust model training and comparison. 

% -----------
%To address the aforementioned limitations, we introduce \textbf{LiveStre4M}, a transformer-based method for real-time streaming of photorealistic novel viewpoints of dynamic scenes from unposed input videos. LiveStre4m combines three key components: a  \textit{Camera Predictor} that estimates camera intrinsics and extrinsics directly from raw RGB images, a \textit{Spatial Module} that performs fast feed-forward 3D Gaussian Splatting for scene reconstruction, and a \textit{Neural Interpolation Network (NIN)} that integrates frame interpolation with super resolution to ensure temporal consistency.

%Inspired by previous methods on pose regression and scene reconstruction \cite{posediff, vggt, FLARE}, the Camera Predictor employs multi-view ViTs to estimate camera parameters without explicit calibration from unposed RGB images. The Spatial module leverages a transformer-based architecture to encode scene geometry and appearance to generate a 3D Gaussian Splatting representation. Although this enables efficient photorealistic NVS, computational costs limit the high-resolution and high-frame-rate streaming. The NIN module alleviates this by using Diffusion Transformers~\cite{DiT} for temporal frame interpolation alongside and a high-speed single-image super-resolution module. enabling temporal consistency and sufficient frame rate and image quality. 
% ----------

To address the aforementioned limitations, we introduce \textbf{LiveStre4m}, a transformer-based method for real-time streaming of photorealistic novel viewpoint videos of dynamic scenes from unposed input videos. LiveStre4m combines three key components: a  \textit{Camera Pose Predictor}, a \textit{Spatial Module}, and a \textit{Neural Interpolation Network (NIN)}. The Camera Pose Predictor, inspired by previous methods on pose regression and scene reconstruction~\cite{posediff, vggt, FLARE}, employs multi-view ViTs to estimate camera parameters directly from unposed RGB images. The Spatial Module leverages a transformer-based architecture to estimate scene geometry and appearance to generate a 3D Gaussian Splatting representation. Although this enables efficient photorealistic NVS, computational costs limit image resolution and video frame rate. The NIN module alleviates this by applying Diffusion Transformers~\cite{DiT} for temporal frame interpolation alongside a high-speed single-image super-resolution module, enabling temporal consistency, sufficient frame rate and improved image resolution.
% ----------

%A large portion of live NVS applications are human-centric, yet existing multi-view datasets are dominated by buildings and indoor scenes~\cite{MegaDepth, ARKitScenes, BlendedMVS, Scannet++}. As a result, pre-trained 3D reconstruction models struggle to generalize to human subjects. To tackle this problem, LiveStre4m is finetuned with multi-view human-centric datasets, including MeetRoom~\cite{meetroom}, ENeRF~\cite{enerf} and N3DV~\cite{cooking_guy}.

LiveStre4m operates end-to-end, taking sparse unposed input video streams and producing novel viewpoint video at $1024 \times 768$ resolution in $0.07$s per frame. To summarize, the main contributions of this work are as follows.

\begin{itemize}
  \item A feed-forward network, \textbf{LiveStre4m}, that enables real-time streaming of photorealistic novel viewpointvideos for dynamic scenes from sparse multi-view video input.
  %\item Efficient camera parameter estimation directly from RGB images using a transformer-based \textit{camera predictor} module. The predicted parameters are used by a \textit{Spatial Module}, capable of efficient zero-shot NVS.  
  \item Efficient zero-shot NVS with a transformer-based \textit{Spatial Module} guided by camera parameters predicted directly from RGB images using a \textit{Camera Pose Predictor}.
  \item \textit{Neural Interpolation Network} module that performs $2\times$ image upscaling and frame interpolation to ensure temporal consistency and image quality.
  \item Comprehensive experiments demonstrating that {LiveStre4m} achieves significantly faster reconstruction than state-of-the-art dynamic NVS methods, enabling live deployment.
\end{itemize}

\section{Related Work}
\label{sec:related_work}

\subsection{Novel View Synthesis}
\label{subsec:nove_view_synthesis}
Recent progress in Novel View Synthesis (NVS) has been largely driven by Neural Radiance Fields (NeRF)~\cite{nerf} and, more recently, by Gaussian Splatting~\cite{3dgs}. Both methods generate unseen viewpoints of a static scene from multiple input images, but they are different in how the scene is formed and rendered.  

NeRF represents a scene by a Multilayer Perceptron (MLP) neural network trained on dense multi-view input images of this scene. The trained network infers density and color for points within the desired view to generate photorealistic novel viewpoints. However, the rendering process is computationally demanding due to the number of network forward passes required. Additionally, the scene-specific overfitting of the MLPs prevents generalization.

Gaussian Splatting takes a different approach by representing the scene explicitly as a set of 3D Gaussians placed in space. These Gaussians are projected (or 'splatted') onto the image plane to render newly generated novel viewpoints. This representation enables much faster rendering and more efficient optimization than NeRF models in static scenes. Despite these improvements, Gaussian Splatting still requires several minutes of per-scene optimization, which limits its usefulness for real-time novel-view streaming.

\subsection{Dynamic 3D Scenes}
\label{subsec:dynamic_3d_scenes}
Beyond static environments, recent work has focused on extending NVS to dynamic scenes. 4D Gaussian Splatting~\cite{4dgs} models motion by inducing time-varying parameters for the Gaussian primitives, enabling novel view rendering at each timestep. However, training a dynamic scene model requires dense coverage multi-view video and multiple optimization iterations over the entire video, making these approaches impractical for live streaming.

To reduce dependency on the entire video, 3DGStream~\cite{3dgstream} proposes a per-frame optimization strategy that incrementally updates a Gaussian model to represent the moving objects as new frames arrive. Although this approach represents a step toward streamable NVS, the optimization requires $12$ seconds per frame, which is far too slow for real-time applications. Building upon this idea, IGS~\cite{igs} is proposed as an even faster approach, reducing reconstruction time to approximately 2.7 seconds per frame, yet this speed still remains insufficient for real-time novel view streaming. 

\subsection{Feed-forward 3D Reconstruction}
\label{subsec:feedforward_3d_reconstruction}
The above methods require dense input view coverage and ground-truth camera poses to accurately reconstruct 3D scenes, which limits their applicability in real-time settings. In contrast, feed-forward 3D reconstruction approaches overcome these issues by predicting scene geometry in a single forward pass.

DUSt3R~\cite{dust3r} introduces a Vision Transformer-based model~\cite{ViT} capable of reconstructing 3D scenes from unposed, sparse input views in one forward pass. DUSt3R is generalizable to unseen scenes, as it is pretrained on large-scale multi-view datasets. MASt3R~\cite{mast3r} further expands this approach, providing more accurate results by adding a local feature-matching head.

These models have served as foundations for feed-forward NVS algorithms~\cite{splatt3r,FLARE,noposplat,anysplat,depthsplat}. Splatt3r~\cite{splatt3r} employs a frozen MASt3R backbone together with a Dense Prediction Transformer~(DPT)~\cite{dpt} head to predict Gaussian parameters for each pixel, enabling zero-shot photorealistic NVS of static scenes from only two input views. FLARE~\cite{FLARE} extends the idea of feed-forward NVS by processing up to 8 input views simultaneously. It employs a two-stage camera prediction and additional appearance modeling to ensure better geometry estimation and visual quality. Although these feed-forward methods operate notably faster than optimization-based approaches, their runtimes remain above the threshold required for live-streaming NVS of 10+ frames per second.  

\subsection{Interpolation}
\label{subsec:interpolation}
Video Frame Interpolation (VFI) focuses on generating synthetic frames between the existing ones. In this field, specific approaches tackle different objectives, such as improving visual quality~\cite{Rangenullspace, TCL, Biformer}, decreasing runtime~\cite{ifrnet, rife}, or handling fast moving objects~\cite{eden, film}. 

Handling fast motion while maintaining short runtimes is crucial to enable live-streaming applications. IFRNet~\cite{ifrnet} introduces a lightweight encoder-decoder network that achieves high visual quality on VFI benchmarks~\cite{vimeo90k,ucf101,snu_film} with minimal runtime, although it is not designed for fast-moving objects. EDEN~\cite{eden} addresses this limitation using a diffusion-based approach, where a diffusion transformer~\cite{DiT} encodes two consecutive frames to generate an intermediate frame. EDEN maintains low latency and achieves reliable performance in videos with fast-moving objects.

\section{Method}
\label{sec:method}
This section presents the proposed LiveStre4m method. An overview of LiveStre4m data flow is first provided, followed by detailed descriptions of the Camera Pose Predictor, Spatial Module, and Neural Interpolation Network (NIN) in~\Cref{subsec:camera_predictor,subsec:spatial_module,subsec:NIN}, respectively. Finally, \cref{subsec:training_strategy} describes the loss function used for finetuning the full model.

\begin{figure*}[t]
  \centering
  \includegraphics[width=\textwidth]{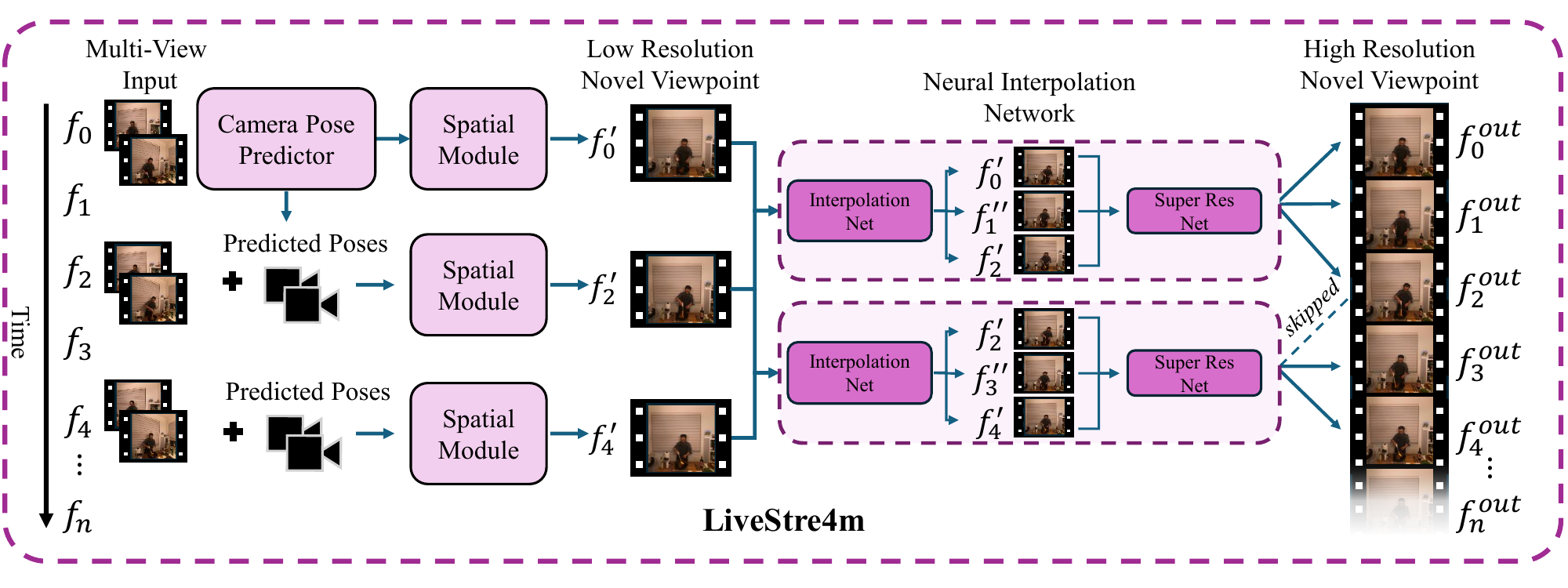}
  \caption{Overviews of LiveStre4m model architecture. The model receives multi-view video keyframes, the first such keyframe is used by the Camera Pose Predictor to regress camera poses. Then, the images and predicted camera information are processed by the Spatial Module to generate a low-resolution novel viewpoint of the scene. After two keyframes are processed, NIN interpolates and increases image resolution to generate the high-resolution video snippet. Snippets are accumulated in a temporally consistent video stream. To accumulate two input keyframe, the generated video has a small delay of less than a second when live-streaming novel viewpoint video.}
  \label{fig:pdf-wide}
\end{figure*}

\subsection{Overview}
State-of-the-art methods for dynamic novel view synthesis (NVS) rely on lengthy optimization and ground-truth camera parameters, which prevents real-time streaming of novel viewpoints. LiveStre4m addresses this limitation with a feed-forward method that enables live streaming of new views for arbitrarily long unposed videos. Moreover, it can efficiently reconstruct and render photorealistic viewpoints in live-streaming applications.

LiveStre4m is composed of three key modules as shown in \cref{fig:livestre4m}. The \textit{Camera Pose Predictor} is capable of predicting camera parameters from unposed input images. The \textit{Spatial Module} (represented as $Sp(\cdot)$ function) generates novel viewpoints from as few as two input views of a scene, and the \textit{Neural Interpolation Network} (NIN) handles the temporal aspect of the novel viewpoint video generation. NIN is further composed of an Interpolation Network (defined as $Inter(\cdot)$ function) that ensures consistency and sufficient frame rate, as well as a Super Resolution (denoted as $SR(\cdot)$ function) lightweight CNN for image upsampling.  

The first step of the proposed method is to predict camera poses from the $n$ unposed input RGB views $f_t \in \mathbb{R}^{n \times 3 \times H \times W}$ where $W, H$ are width and height of the images in pixels, respectively. The Camera Pose Predictor module generates the camera extrinsic and intrinsic matrices as well as the camera embedding for each input view. This camera information is used to guide further scene reconstruction.  

At a timestep $t$, the multi-view frame $f_t$ and the predicted camera parameters are processed by $Sp(\cdot)$, resulting in an arbitrary number $m$ of novel viewpoints of the scene. Then, $f_t'$ is accumulated until $f_{t+2}'$ is processed. Both multi-view frames are processed by $Inter(\cdot)$ to generate an intermediate frame $f_{t+1}'' \in \mathbb{R}^{m \times 3 \times H \times W}$ from the same viewpoints. Finally, the three frames are fed into $SR(\cdot)$ to generate $f_t^{\text{out}}$, $f_{t+1}^{\text{out}}$ and $f_{t+2}^{\text{out}} \in \mathbb{R}^{m \times 3 \times (H\times2) \times (W\times2)}$

\begin{equation}
\forall t \in \{0, 2, 4, \dots, n\!-\!2\},
\label{eq:model_1}
\end{equation}
\begin{equation}
    f_t' = Sp(f_t), \quad f_{t+2}'  = Sp(f_{t+2}),  
    \label{eq:model_2}
\end{equation}
\begin{equation}
 f_t', f_{t+1}'', f_{t+2}' = Inter(f_t', f_{t+2}'),
 \label{eq:model_3}
 \end{equation}
 \begin{equation}
     f_t^{\text{out}}, f_{t+1}^{\text{out}}, f_{t+2}^{\text{out}} = SR(f_t', f_{t+1}'', f_{t+2}')
 \label{eq:model_4}
\end{equation}

\Cref{eq:model_1,eq:model_2,eq:model_3,eq:model_4} are the mathematical formulations of the data flow through the proposed Spatial Module and the NIN. The following subsections provide a detailed explanation of each key component of the LiveStre4m model.

\subsection{Camera Pose Predictor}
\label{subsec:camera_predictor}
Camera pose prediction is an inherently challenging task, particularly in the absence of ground-truth camera parameters. To address this, \textit{Camera Pose Predictor} is proposed with a coarse-to-fine strategy, following the implementation proposed by FLARE~\cite{FLARE}.

Coarse poses are first estimated by processing tokenized input frames $f_0$ concatenated with learnable camera tokens through a transformer decoder. The resulting coarse camera pose is refined by another transformer module that extracts local view-centric information. Finally, an attention head outputs the fine-grained camera parameters. 

The camera intrinsics are determined under the assumption that the principal points $(c_x, c_y)$ are centered. The focal lengths in pixels are computed as $focal_x = focal \cdot W$ and $focal_y = focal \cdot H$, where $W$ and $H$ denote image width and height, respectively. The estimated camera parameters for each view consist of an extrinsic matrix and an intrinsic matrix represented by

\begin{equation}
    \text{extrinsics} = \begin{bmatrix}
         R &  T
    \end{bmatrix}, R \in \mathbb{R}^{3 \times 3}, T \in \mathbb{R}^{3} , 
    \label{eq:extrinsic}
\end{equation}
\begin{equation}
        \text{intrinsics} = \begin{bmatrix} focal_x & focal_y & c_x & c_y\end{bmatrix}^T .
        \label{eq:intrinsic}
\end{equation}

The refined camera pose embedding $P_{fine}$ is composed of the predicted rotation quaternion $\in \mathbb{R}^{n \times 4}$ and the normalized translation vector $\in \mathbb{R}^{n \times 3}$,  producing a vector of $P_{fine} \in \mathbb{R}^{n \times 7}$. The predicted camera parameters are used to guide the 3D scene reconstruction in the Spatial Module. 

In this work, cameras are considered static throughout the entire duration of the multi-view video, therefore, the camera parameters are estimated only at $t=0$. Although the proposed approach can generalize to dynamic  cameras, this would require camera prediction at every input frame $f_t \quad \forall t \in \{0, 2, 4, \dots, n\!-\!2\}$, significantly increasing runtime and impeding live-streaming applications. Therefore, moving camera scenarios are not explored in this work to focus on live-streaming NVS.

\begin{figure*}[ht]
  \centering
  \includegraphics[width=\textwidth]{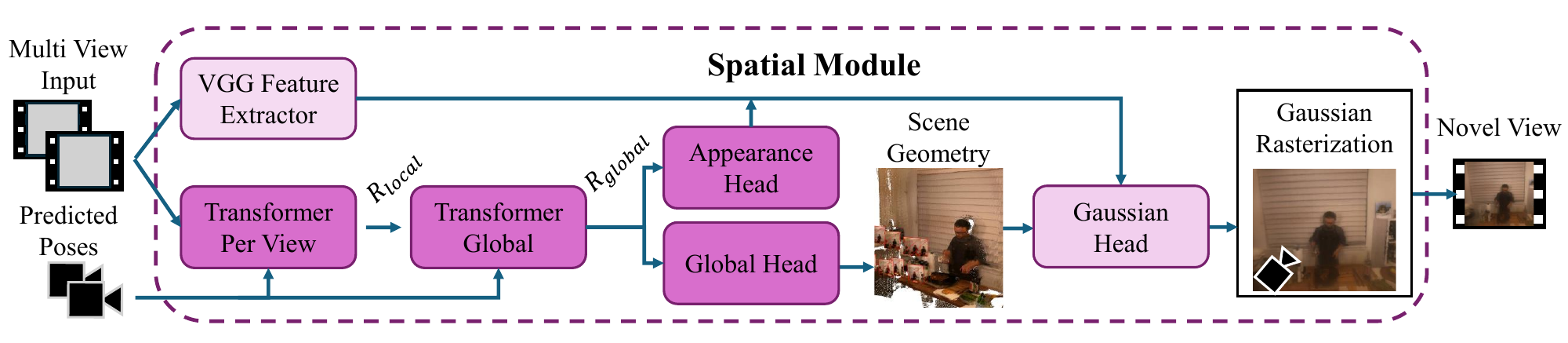}
  \caption{Representation of the Spatial Module architecture. The module leverages ViTs to model scene 3D geometry and appearance as well as a VGG model to extract visual features. By combining appearance features extracted and the predicted geometry, a DPT head predicts 3D Gaussian parameters centered on the predicted pointmap to achieve photorealistic NVS.}
  \label{fig:spatial}
\end{figure*}

\subsection{Spatial Module}
\label{subsec:spatial_module}
The Spatial Module is responsible for reconstructing the scene geometry and rendering photorealistic novel viewpoints. An overview of this module is depicted in~\cref{fig:spatial}. At time step $t$, visual features are extracted from the multi-view input $f_t$ using a frozen VGG network~\cite{vgg}. The images are then tokenized and combined with the predicted camera parameters $P_{fine}$, obtained from the \textit{Camera Pose Predictor} module, to create camera-centric representations. These tokens are subsequently processed by the transformer module $\mathcal{F}_{loc}(\cdot)$, which generates a camera-centric representation of the 3D scene relative to the viewing pose, as described by the following equation:

\begin{equation}    
R_{local} = \mathcal{F}_{loc}(f_t, P_{fine}).
\end{equation}

These camera-centric representations serve as geometry priors for estimating the global scene geometry. To enhance robustness against minor inaccuracies in camera pose predictions, explicit geometric projections are avoided~\cite{FLARE}. Instead, a transformer module $\mathcal{F}_{global}(\cdot)$ learns to map local point tokens into a single global scene token vector, guided by the predicted camera poses. A deep representation of the global scene geometry $R_{global}$ is obtained by:

\begin{equation}
R_{global} = \mathcal{F}_{global}(R_{local}, P_{fine}).
\end{equation}

The global scene geometry vector $R_{global}$ is upsampled by the Dense Prediction Transformer~(DPT) Global Head to form a dense 3D point map. Although point maps provide an explicit and efficient representation of 3D scenes, they lack sufficient visual detail for photorealistic NVS. In contrast, 3D Gaussian representations have demonstrated better photorealism in NVS~\cite{3dgs}. Accordingly, we initialize Gaussian primitives at point map coordinates and predict their appearance-based parameters: spherical harmonics, opacity, rotation, and scale.

A DPT Appearance Head learns to extract an appearance latent vector, which is then fused with the features extracted by the VGG network. The resulting vector is processed by a CNN Gaussian Head that regresses the remaining Gaussian parameters, yielding a complete 3D Gaussian scene representation.

Finally, fully differentiable Gaussian rasterization~\cite{3dgs} is applied to generate novel viewpoints of the scene. While the number of generated viewpoints has only a minor impact on runtime, an increased number of generated views can hinder real-time streaming. This effect is explored in detail in~\cref{subsec:ablation_studies}. Ultimately, the Spatial Module outputs renderings of the reconstructed scene at time $t$ from selected novel viewpoints.

\subsection{Neural Interpolation Network}
\label{subsec:NIN}
The Spatial Module, composed of multiple transformers, demands high computational power. To ensure low latency in live-streaming applications, this module is restricted to reconstruction of low-resolution keyframes. Furthermore, since the Spatial Module operates on each timestamp independently, it does not guarantee temporal consistency across frames. 

To address these challenges, the Neural Interpolation Network (NIN) is introduced, following the implementation of EDEN~\cite{eden} pretrained for large-motion frame interpolation. It employs a Diffusion Transformer~(DiT) guided by an encoded vector representing the difference between two keyframes, $f_t$ and $f_{t+2}$. Through this mechanism, NIN generates an intermediate frame $f_{t+1}$, thereby enforcing temporal consistency between frames.

Finally, all three frames ($f_t$, $f_{t+1}$ and $f_{t+2}$) are fed into the Super Resolution module $SR(\cdot)$ to enhance image resolution. This module is a lightweight pretrained CNN architecture~\cite{sr} performing fast image super-resolution. The resulting high-resolution frames are used to compose the novel view video stream of LiveStre4m.

It is important to note that, as shown in \cref{fig:livestre4m}, from the second iteration of the NIN onward, the first generated frame is discarded. This step is essential for maintaining temporal consistency, ensuring that the temporal offset between consecutive frames in the generated video is the same. Furthermore, this allows LiveStre4m to be applied to various input frame rates while preserving temporal consistency in the generated novel viewpoint video.

\begin{table*}[t]
\centering
\caption{Quantitative comparison of dynamic scene reconstruction methods on Neural3DVideo~\cite{cooking_guy} and MeetRoom~\cite{meetroom} datasets at $1352 \times 1014$ and $ 1280 \times 720$ resolution, respectively. Reported metrics include average runtime, PSNR, number of input views, and whether methods are camera-free. For a fair comparison, LiveStre4m is evaluated on the same A100 GPU as reported by IGS.}
\resizebox{\textwidth}{!}{%
\begin{tabular}{l | c c c c | c c c c}
\toprule
\multirow{2}{*}{\textbf{Model}} &
\multicolumn{4}{c|}{\textbf{Neural3DVideo~\cite{cooking_guy}}} &
\multicolumn{4}{c}{\textbf{MeetRoom~\cite{meetroom}}} \\
\cmidrule(lr){2-5} \cmidrule(lr){6-9}
 & Runtime (s) $\downarrow$ & PSNR $\uparrow$ & Camera Free & \#Views $\downarrow$ 
 & Runtime (s) $\downarrow$ & PSNR $\uparrow$ & Camera Free & \#Views $\downarrow$ \\
\midrule
\multicolumn{9}{l}{\textit{Video Optimization}} \\
K-planes~\cite{kplanes}           & 48.00 & 32.17 & \ding{55} & $\ge$19 & -- & -- & -- & -- \\
4DGS~\cite{4dgs}              & 7.80 & 32.70 & \ding{55} & $\ge$19 & -- & -- & -- & -- \\
Spacetime-GS~\cite{spacetime_gauss}     & 48.00 & 33.71 & \ding{55} & $\ge$19 & -- & -- & -- & -- \\
\midrule
\multicolumn{9}{l}{\textit{Frame Optimization}} \\
StreamRF~\cite{meetroom}          & 15.00 & 32.09 & \ding{55} & $\ge$19 & -- & -- & -- & -- \\
3DGStream~\cite{3dgstream}           & 16.93 & 32.75 & \ding{55} & $\ge$19 & 11.51 & 28.36 & \ding{55} & 12 \\
IGS~\cite{igs}                & 2.67 & 33.89 & \ding{55} & $\ge$19 & 2.77 & 29.24 & \ding{55} & 12 \\
\midrule
\multicolumn{9}{l}{\textit{Feed-forward}} \\
LiveStre4m (Ours)   & \textbf{0.14} & 20.64  & \ding{51} & \textbf{2} & \textbf{0.10} & 18.17 & \ding{51} & \textbf{2} \\
\bottomrule
\end{tabular}%
}
\label{tab:combined_results}
\end{table*}

\subsection{Training Strategy}
\label{subsec:training_strategy}

The proposed method is finetuned with visual losses only, \textit{i.e.}, using only pixel information from video frames without requiring ground-truth camera poses or additional scene data. The overall loss function is defined by 

\begin{equation}
    \mathcal{L} = \lambda_{MSE} \, \|\hat f_t^{\text{out}} - f_t^{\text{out}}\|_2^2
+ \lambda_{L} \,\text{LPIPS}(\hat f_t^{\text{out}}, f_t^{\text{out}}),
\label{eq:loss}
\end{equation}
where $\lambda_{MSE}$ and $\lambda_{L}$ are hyperparameters and $LPIPS$ is a perceptual metric \cite{lpips}.
 
Since LiveStre4m employs multiple Transformers, the computational cost to train the full model is substantial. To facilitate training, model weights are initialized from pretrained components, and then the entire architecture is finetuned for a limited number of  epochs. Further details are given in the following section.  

%Camera Predictor is initialized using pretrained modules from FLARE \cite{FLARE} and new linear adapters, for camera rotation and translation and focals. To train this module, we used it to predict camera information from unposed input views and used the predicted poses to guide the NVS synthesis process of the pre-trained Spatial Module. The loss is calculated between the generated novel views and the input images. During this process, we also fine-tuned the Spatial Module with smaller updates to ensure consistency between predicted camera poses while still leveraging pretrained weights.

%The NIN module is finetuned by processing two downscaled frames, $f_t$ and $f_{t+2}$ to obtain the upscaled outputs $f_t^{\text{out}}$, $f_{t+1}^{\text{out}}$ and $f_{t+2}^{\text{out}}$, which are compared to the corresponding ground-truth frames.

\section{Experiments}
\label{sec:experiments}

% TABLE OF RESULTS IN METHODS FOR BETTER VISUALIZATION ON THE PAPER PDF

\subsection{Datasets}
\label{subsec:datasets}
Two widely used datasets are employed for experimental evaluation. The first one being the Neural3DVideo~\cite{cooking_guy} dataset, which contains six dynamic scenes captured by a multi-view setup with 21 cameras at a resolution of 2704×2028. Each multi-view video consists of 300 frames. The second dataset is the MeetRoom~\cite{meetroom}, comprising three dynamic scenes recorded with 13 cameras at a resolution of 1280 × 720. Similarly, each multi-view video contains 300 frames. In each dataset, one camera view is reserved as the target viewpoint. Following the training strategy proposed by IGS~\cite{igs}, LiveStre4m is finetuned using four scenes of Neural3DVideo~\cite{cooking_guy} dataset, while the remaining two scenes (\textit{cut\_roasted\_beef} and \textit{sear\_steak}) serve as the test set, notably, the latter lacks one input view. Additionally, the MeetRoom dataset~\cite{meetroom} is not used for finetuning, but only for testing, as it serves for evaluation of the generalization capability. 

\subsection{Implementation Details}
\label{subsec:implementation_details}
The Spatial Module is composed of a ViT backbone comprising one encoder with 24 blocks and two decoders, each consisting of 12 blocks of 768 dimensional embeddings. Downstream DPT heads leverage the extracted features to reconstruct the scene geometry. The Camera Pose Predictor consists of the same pretrained ViT backbone as the Spatial Module together with a two-stage pose predictor composed of small attention blocks and MLPs. The interpolation component of the NIN is a pretrained 12-block DiT with spatial and temporal attention coupled with a 4-block decoder that refines image tokens to generate the interpolated frame. The super-resolution module is a lightweight 3 layer CNN with 32 channels pretrained for 2$\times$ image upsampling.

The finetuning is conducted on a single H100 GPU using the scenes described in~\Cref{subsec:datasets}. The Camera Pose Predictor and Spatial Module are finetuned for three iterations across all frames of the input videos. Subsequently, the full LiveStre4m architecture, containing 2B parameters, is optimized for five iterations on the first 50-frame segments of the same videos. 

%\subsection{Inference}
%\label{subsec:inference}
%During inference, given two or more input streams, the Camera Predictor estimates camera poses and focal lengths from the first multi-view input. The Spatial Module then synthesizes the output keyframes for the target viewpoints, using the predicted camera parameters and input keyframes. These keyframes are processed by the NIN to achieve high-resolution frames. Finally, LiveStre4m accumulates all generated frames to compose the continuous novel-view video stream.

\subsection{Results}
\label{subsec:results}

\begin{table*}[t]
\centering
\caption{Quantitative comparison of feed-forward scene reconstruction methods on Neural3DVideo~\cite{cooking_guy} and MeetRoom~\cite{meetroom} datasets on a single H100 GPU}
\resizebox{\textwidth}{!}{%
\begin{tabular}{l | c c c c c | c c c c c}
\toprule
\multirow{2}{*}{\textbf{Model}} &
\multicolumn{5}{c|}{\textbf{Neural3DVideo~\cite{cooking_guy}}} &
\multicolumn{5}{c}{\textbf{MeetRoom~\cite{meetroom}}} \\
\cmidrule(lr){2-6} \cmidrule(lr){7-11}
 & Runtime (s) $\downarrow$ & PSNR $\uparrow$ & Resolution $\uparrow$ & Pose Free & \#Views $\downarrow$ 
 & Runtime (s) $\downarrow$ & PSNR $\uparrow$ & Resolution $\uparrow$ & Pose Free & \#Views $\downarrow$ \\
\midrule
%\multicolumn{9}{l}{\textit{Feed-forward}} \\
FLARE~\cite{FLARE}  & 0.249 & 21.45 & $512 \times 384$ & \ding{51} & 2 & 0.243 & 16.65 &  $512 \times 384$ & \ding{51} & 2 \\
LiveStre4m (Ours)   &  0.074 & 21.11  &  $1024 \times 768$ & \ding{51} & 2 & 0.074 & 18.65 &  $1024 \times 768$ & \ding{51} & 2 \\
LiveStre4m (Ours)   &  \textbf{0.062} & 22.44  &  $512 \times 384$  & \ding{51} & 2 & \textbf{0.062} & 19.32 &   $512 \times 384$ & \ding{51} & 2 \\
\bottomrule

\end{tabular}%
}
\label{tab:combined_results_feedforward}
\end{table*}

In this subsection, experimental results are presented, including model runtime, and PSNR, to assess NVS quality. Additionally, the number of input views and the requirement for camera information are considered to benchmark the proposed approach against the state-of-the-art. For consistent reporting, runtime is defined as the time required to generate a novel viewpoint of a complete video, averaged by the number of frames, excluding pre-processing. To ensure fair comparison with methods that rely on ground-truth camera parameters, we follow the evaluation protocol established by prior pose-free approaches~\cite{noposplat,instantsplat,nerfmm}.

Table~\ref{tab:combined_results} presents a detailed quantitative comparison of dynamic scene reconstruction methods evaluated on the Neural3DVideo~\cite{cooking_guy} and MeetRoom~\cite{meetroom} datasets. State-of-the-art methods are grouped into three categories: \textit{video-optimization}, \textit{frame-optimization}, and \textit{feed-forward} approaches. The results in the table, except those of the LiveStre4m model, are taken from the IGS~\cite{igs} paper. While both video-optimization and frame-optimization methods use over 19 input views for the Neural3DVideo dataset, LiveStre4m operates with only two views, representing a substantial reduction in the number of required input viewpoints. Similarly, frame-optimization methods require 12 input views on the MeetRoom dataset, whereas LiveStre4m again needs only two. Moreover, unlike video-optimization and frame-optimization methods that rely on known camera parameters, LiveStre4m performs novel view synthesis in a fully pose-free manner. 

In terms of computational efficiency, the best-performing video-optimization method, 4DGS, achieves a runtime of 7.8 seconds per frame, while the fastest frame-optimization approach, IGS, achieves 2.67 seconds. It should be noted that the IGS results reported in the table correspond to \textit{IGS-s} model, the fastest version in terms of runtime proposed in their paper~\cite{igs}. In contrast, LiveStre4m operates in 0.14 seconds per frame, approximately $55\times$ faster than 4DGS and $19\times$ faster than IGS, demonstrating a remarkable improvement in efficiency. At the same time, we must note that the resulting PSNR scores are significantly below state-of-the-art and require additional enhancements, outlined in \cref{subsec:enhancements}. In other words, while the optimization-based methods achieve higher visual quality, LiveStre4m demonstrates a significant advantage in streamability, operating substantially faster without requiring prior camera information. 
\\
\begin{figure}[ht]
    \centering
    \includegraphics[width=\columnwidth]{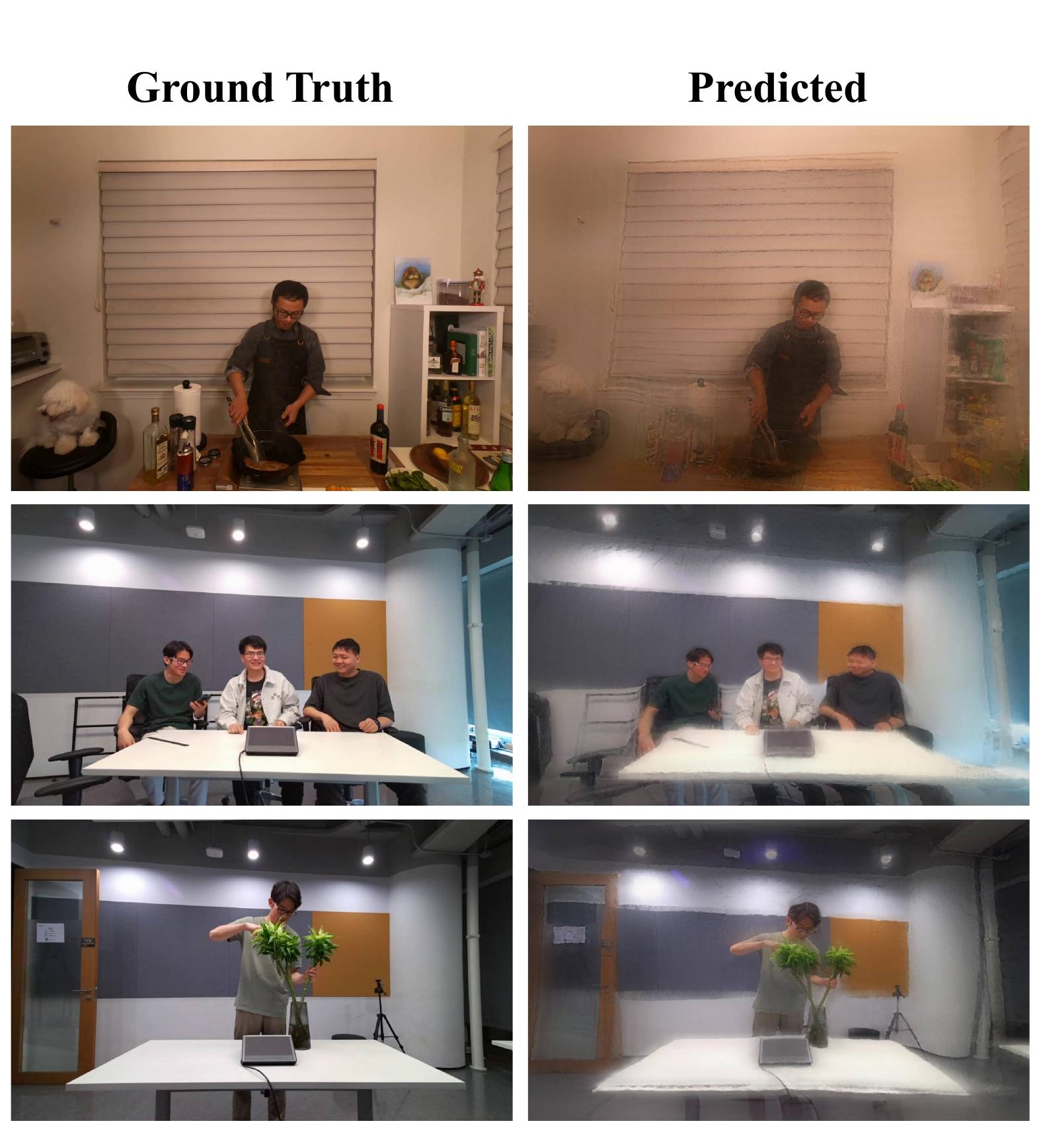}
    \caption{Qualitative results produced by with LiveStre4m, which synthesizes the target viewpoint using only two neighboring input views, without requiring optimization or ground-truth camera parameters. All images are shown at the same resolution as in~\cref{tab:combined_results} From top to bottom: \textit{sear\_steak}~\cite{cooking_guy}, \textit{discussion}~\cite{meetroom} and \textit{trimming}~\cite{meetroom} scene results are provided.}
    \label{fig:quali}
\end{figure}
\\

To further evaluate performance, LiveStre4m is compared with another feed-forward camera-free method FLARE~\cite{FLARE}. Although FLARE is not developed for dynamic scenes, it is capable of zero-shot camera-free NVS from sparse input views. For comparison, FLARE is evaluated with the same pipeline as LiveStre4m at each each timestep and its performance is averaged for the entire video. In this comparison, both methods receive two input view streams closest to the target view and generate 300 output frames from the target viewpoint.

As shown in~\Cref{tab:combined_results_feedforward}, LiveStre4m achieves competitive rendering quality compared to state-of-the-art methods. On the Neural3DVideo dataset, LiveStre4m attains a PSNR of 21.11, closely matching FLARE (21.45).  On the MeetRoom dataset, it outperforms FLARE, achieving 18.65 compared to 16.65. In terms of efficiency, LiveStre4m demonstrates a substantial advantage with a runtime of 0.074s per frame, over $3.3\times$ faster than FLARE (0.245s). Moreover, LiveStre4m performs NVS at twice the output image resolution due to the Super Resolution module. % LiveStre4m achieves comparable visual quality while significantly outperforming FLARE in runtime efficiency. 

Qualitative results are illustrated in~\cref{fig:quali}, comparing reference renderings with LiveStre4m outputs across different test scenes, showcasing photorealistic feed-forward 3D scene reconstruction. We present additional results in supplementary material, including detailed per-scene analysis~(\cref{sec:App}), an evaluation on the highly dynamic VRU-Basketball~\cite{vru_data1,vru_data2} dataset in~(\cref{sec:dynamic_data}), pose prediction accuracy~(\cref{sec:pose}), and the discussion of real-world deployment~(\cref{sec:deployment}).

\subsection{Ablation Studies}
\label{subsec:ablation_studies}
Several ablation studies are conducted to evaluate the performance of the proposed method under different scenarios.  

\textbf{Image Resolution:} \Cref{tab:ablation_res} explores the effect of output image resolution on runtime and visual quality in NVS. Higher resolutions increase computational demand, resulting in longer runtimes. Therefore, a trade-off emerges between visual quality and streamability. As shown in \cref{tab:ablation_res}, maintaining frame rates above $>10$ fps, requires limiting the image resolution to $1152 \times 768$. 

Interestingly, PSNR declines at the higher image resolutions. This can be attributed to the data employed for pretraining multiple LiveStre4m modules which is predominantly lower-resolution images~\cite{sr,FLARE,eden}.  

\begin{table}[ht]
\caption{LiveStre4m performance comparison using different image resolutions, with the same 2 input views and target view, computed on a single H100 GPU.}
\centering
\resizebox{0.6\columnwidth}{!}{
\begin{tabular}{l|cccc}
\toprule
 Output Res. (p) & Runtime (s)$\downarrow$ & PSNR $\uparrow$  \\ 
\midrule
 $512 \times 384$         &    0.062  & 20.57 \\
 $640 \times 512$         &    0.063  & 20.93         \\
 $1024 \times 768$         &   0.074  & 19.63        \\
 $1152 \times 768$        &    0.078  & 19.54          \\
 $1536 \times 1152$       &    0.124  & 17.34         \\
\bottomrule
\end{tabular}}
\label{tab:ablation_res}
\end{table}

\textbf{View Distance and Scene Coverage:} To investigate the influence of input camera placement, we study both the distance between input and target cameras and the number of input views. Input views of Neural3DVideo are arranged in a semi-circle and we grouped them into three categories based on inter-camera distance: the four nearest views (\textit{Closest}), the next four (\textit{Intermediate}), and the four most distant views (\textit{Distant}). As shown in~\Cref{tab:ablation_views}, \textit{Closest} and \textit{Intermediate} views consistently yield higher visual quality than \textit{Distant} views, due to increased overlap with the target view. Furthermore, increasing the number of input views does not improve PSNR. With more inputs, the number of 3D correspondences grows, causing small matching errors to accumulate and degrade reconstruction quality. 

%\begin{figure}[h]
%    \centering
%    \includegraphics[width=0.8\columnwidth]{sec/figures/camera_dist.pdf}
%    \caption{Caption}
%    \label{fig:placeholder}
%\end{figure}

\begin{table}[b]
\caption{Performance of LiveStre4m under varying input view configurations, considering both distance to the target view and the total number of input views, evaluated at $1024 \times 768$ resolution. Due to the camera distribution in the MeetRoom dataset~\cite{meetroom}, consistent categorization into closest, intermediate, and distant views is not feasible. Therefore, results are reported only on the Neural3DVideo dataset~\cite{cooking_guy}.}
\centering
\resizebox{0.7\columnwidth}{!}{%
\begin{tabular}{l|cccc}
\toprule
 Distance    & \# Inputs  & Runtime (s)$\downarrow$ & PSNR $\uparrow$  \\ \midrule
\multirow{2}{*}{\textit{Closest}}      &   2  &    0.074   &  21.11         \\
                              &   4  &    0.085   &  20.82      \\\midrule
\multirow{2}{*}{\textit{Intermediate}} &   2  &    0.074   &  20.24     \\
                              &   4  &    0.085   &  20.05     \\\midrule
\multirow{2}{*}{\textit{Distant}}     &    2  &    0.074   &  18.83     \\ 
                              &   4  &    0.084   &  18.02     \\
\bottomrule

\end{tabular}%
}
\label{tab:ablation_views}
\end{table}

\textbf{Number of Output Views:} As discussed in~\cref{subsec:spatial_module}, generating more output views increases the computation complexity. \Cref{tab:ablation_num_views} explores how the number of output views affects the runtime and NVS quality. Results indicate that increasing the number of output images not only increases the runtime, but also reduces image quality, as target viewpoints deviate further from the central viewpoint.

\begin{table}[ht]
\caption{Ablation study on the number of synthetic views generated by LiveStre4m at $1024 \times 768$ resolution using a single H100 GPU.}
\centering
\resizebox{0.55\columnwidth}{!}{%
\begin{tabular}{c|cccc}
\toprule
 \# Outputs  & Runtime (s)$\downarrow$ & PSNR $\uparrow$  \\ \midrule
 1     &      0.075         & 19.63                    \\
 2     &      0.079         & 19.24               \\
 4     &      0.088         & 18.90                 \\\bottomrule

\end{tabular}%
}
\label{tab:ablation_num_views}
\end{table}

\textbf{Runtime of Model Components:} \Cref{tab:ablation_timers} breaks down the computational cost of each component, highlighting potential bottlenecks. The Spatial Module dominates the runtime with $52.1$ ms. The Interpolation Net is significantly lighter at $19.3$ ms, validating the choice of interpolating intermediate frames for a more efficient model.

\begin{table}[ht]
\centering
\caption{Average runtime of each model component over a 300-frame video at $1024 \times 768$ resolution, measured on a single NVIDIA H100 GPU.}
\label{tab:ablation_timers}
\resizebox{0.6\columnwidth}{!}{
\begin{tabular}{l|S[table-format=2.1]}
\hline
Component             & \multicolumn{1}{c}{Runtime (ms)} \\ \hline
Camera Pose Predictor & 1.5                              \\
Spatial Module        & 52.1                             \\
Gaussian Rendering    & 9.6                              \\
Interpolation Net     & 19.3                             \\
Super Res Net         & 0.6                              \\ \hline
\end{tabular}%
}
\end{table}

\subsection{Discussion}
\label{subsec:enhancements}
LiveStre4m prioritizes efficiency, trading some visual quality for significantly faster inference compared to optimization-based approaches, but to close this gap, potential enhancements are on exploration. Incorporating~\textit{online} pose refinement and lightweight  bundle adjustment could improve geometric consistency and visual fidelity. Furthermore, scaling multi-view video training data is expected to further enhance generalization and output quality.

\section{Conclusion}
\label{sec:conclusion}
This paper introduces LiveStre4m, a zero-shot, camera-free method for live streaming of novel-viewpoint video from sparse input views. LiveStre4m consists of the \textit{Camera Pose Predictor}, \textit{Spatial Module}, and \textit{Neural Interpolation Network}~(NIN). The Camera Pose Predictor estimates camera poses from sparse unposed views, which are then used by the Spatial Module, composed of multi-view ViTs, to reconstruct the scene and generate photorealistic novel viewpoints. NIN ensures temporal consistency, better image resolution and higher frame rates. 

%the \textit{Camera Pose Predictor}, capable of predicting camera poses from unposed views, \textit{Spatial Module}, for photorealistic NVS and finally, \textit{NIN}, which handles temporal dependency of live-streaming NVS.

%

% the \textit{Camera Pose Predictor}, capable of predicting camera poses from unposed views, \textit{Spatial Module}, for photorealistic NVS and finally, \textit{NIN}, which handles temporal dependency of live-streaming NVS

%We introduce the Camera Predictor module, capable of predicting camera poses from sparse unposed views, a Spatial module, composed of multi-view ViTs for photorealistic scene reconstruction and NVS and finally, the NIN, that ensures temporal consistency, improved image resolution and frame-rate. 

Experimental results demonstrate that the proposed method produces photorealistic novel viewpoints from sparse unposed input views at $1024 \times 768$ resolution in only $0.07s$ per frame on a single GPU. Although LiveStre4m does not outperform optimization-based methods in visual quality, its runtime efficiency enables practical live-streaming applications. Furthermore, it achieves comparable PSNR to  the state-of-the-art in feed-forward camera-free NVS, while operating significantly faster. Overall, LiveStre4m represents a significant step toward live-streaming of novel-viewpoint video in real-time.

% \section*{Acknowledgements}
% \raggedright
% This work is financially supported by the 2023022 ELEVATION Xecs-Eureka project.
\section*{Acknowledgments}
This work was supported by the ELEVATION Xecs 2023022 project 

{
    \small
    \bibliographystyle{ieeenat_fullname}
    \bibliography{main}

@String(IJCV = {Int. J. Comput. Vis.})

@String(CVPR= {IEEE Conf. Comput. Vis. Pattern Recog.})

@String(ECCV= {Eur. Conf. Comput. Vis.})

@String(TOG= {ACM Trans. Graph.})

@String(ICLR = {Int. Conf. Learn. Represent.})

@String(AAAI = {AAAI})

@String(IJCV  = {IJCV})

@String(CVPR  = {CVPR})

@String(ECCV  = {ECCV})

@String(TOG   = {ACM TOG})

@String(ICLR  = {ICLR})

@Article{3dgs,
      author       = {Kerbl, Bernhard and Kopanas, Georgios and Leimk{\"u}hler, Thomas and Drettakis, George},
      title        = {3D Gaussian Splatting for Real-Time Radiance Field Rendering},
      journal      = {ACM Transactions on Graphics},
      number       = {4},
      volume       = {42},
      month        = {July},
      year         = {2023},
      url          = {https://repo-sam.inria.fr/fungraph/3d-gaussian-splatting/}
}

@InProceedings{4dgs,
                        author    = {Wu, Guanjun and Yi, Taoran and Fang, Jiemin and Xie, Lingxi and Zhang, Xiaopeng and Wei, Wei and Liu, Wenyu and Tian, Qi and Wang, Xinggang},
                        title     = {4D Gaussian Splatting for Real-Time Dynamic Scene Rendering},
                        booktitle = {Proceedings of the IEEE/CVF Conference on Computer Vision and Pattern Recognition (CVPR)},
                        month     = {June},
                        year      = {2024},
                        pages     = {20310-20320}
                    }

@inproceedings{4dgauss,
  title={Real-time Photorealistic Dynamic Scene Representation and Rendering with 4D Gaussian Splatting},
  author={Yang, Zeyu and Yang, Hongye and Pan, Zijie and Zhang, Li},
  booktitle={International Conference on Learning Representations (ICLR)},
  year={2024}
}

@InProceedings{spacetime_gauss,
    author    = {Li, Zhan and Chen, Zhang and Li, Zhong and Xu, Yi},
    title     = {Spacetime Gaussian Feature Splatting for Real-Time Dynamic View Synthesis},
    booktitle = {Proceedings of the IEEE/CVF Conference on Computer Vision and Pattern Recognition (CVPR)},
    month     = {June},
    year      = {2024},
    pages     = {8508-8520}
}

@inproceedings{3dgstream,
  title={3DGStream: On-the-Fly Training of 3D Gaussians for Efficient Streaming of Photo-Realistic Free-Viewpoint Videos},
  author={Jiakai Sun and Han Jiao and  Guangyuan Li and Zhanjie Zhang and Lei Zhao and Wei Xing},
  booktitle={CVPR},
  year={2024}
}

@inproceedings{dust3r,
      title={DUSt3R: Geometric 3D Vision Made Easy}, 
      author={Shuzhe Wang and Vincent Leroy and Yohann Cabon and Boris Chidlovskii and Jerome Revaud},
      booktitle = {CVPR},
      year = {2024}
}

@misc{mast3r,
      title={Grounding Image Matching in 3D with MASt3R}, 
      author={Vincent Leroy and Yohann Cabon and Jerome Revaud},
      booktitle = {ECCV},
      year = {2024}
}

@article{splatt3r,
  title={Splatt3r: Zero-shot gaussian splatting from uncalibrated image pairs},
  author={Smart, Brandon and Zheng, Chuanxia and Laina, Iro and Prisacariu, Victor Adrian},
  journal={arXiv preprint arXiv:2408.13912},
  year={2024}
}

@inproceedings{FLARE,
  title={Flare: Feed-forward geometry, appearance and camera estimation from uncalibrated sparse views},
  author={Zhang, Shangzhan and Wang, Jianyuan and Xu, Yinghao and Xue, Nan and Rupprecht, Christian and Zhou, Xiaowei and Shen, Yujun and Wetzstein, Gordon},
  booktitle={Proceedings of the Computer Vision and Pattern Recognition Conference},
  pages={21936--21947},
  year={2025}
}

@inproceedings{pixelsplat,
      title={pixelSplat: 3D Gaussian Splats from Image Pairs for Scalable Generalizable 3D Reconstruction},
      author={David Charatan and Sizhe Li and Andrea Tagliasacchi and Vincent Sitzmann},
      year={2024},
      booktitle={CVPR},
}

@article{quark,
  author = {Flynn, John and Broxton, Michael and Murmann, Lukas and Chai, Lucy and DuVall, Matthew and Godard, Cl\'{e}ment and Heal, Kathryn and Kaza, Srinivas and Lombardi, Stephen and Luo, Xuan and Achar, Supreeth and Prabhu, Kira and Sun, Tiancheng and Tsai, Lynn and Overbeck, Ryan},
  title = {Quark: Real-time, High-resolution, and General Neural View Synthesis},
  year = {2024},
  issue_date = {December 2024},
  publisher = {Association for Computing Machinery},
  address = {New York, NY, USA},
  volume = {43},
  number = {6},
  issn = {0730-0301},
  url = {https://doi.org/10.1145/3687953},
  doi = {10.1145/3687953},
  journal = {ACM Trans. Graph.},
  month = nov,
  articleno = {194},
  numpages = {20}
}

@inproceedings{cooking_guy,
   author = {Tianye Li and Mira Slavcheva and Michael Zollhoefer and Simon Green and Christoph Lassner and Changil Kim and Tanner Schmidt and Steven Lovegrove and Michael Goesele and Richard Newcombe and Zhaoyang Lv},
   doi = {10.1109/CVPR52688.2022.00544},
   isbn = {9781665469463},
   issn = {10636919},
   booktitle = {Proceedings of the IEEE Computer Society Conference on Computer Vision and Pattern Recognition},
   pages = {5511-5521},
   publisher = {IEEE Computer Society},
   title = {Neural 3D Video Synthesis from Multi-view Video},
   volume = {2022-June},
   year = {2022}
}

@article{meetroom,
  title={Streaming radiance fields for 3d video synthesis},
  author={Li, Lingzhi and Shen, Zhen and Wang, Zhongshu and Shen, Li and Tan, Ping},
  journal={Advances in Neural Information Processing Systems},
  volume={35},
  pages={13485--13498},
  year={2022}
}

@misc{nerf,
      title={NeRF: Representing Scenes as Neural Radiance Fields for View Synthesis}, 
      author={Ben Mildenhall and Pratul P. Srinivasan and Matthew Tancik and Jonathan T. Barron and Ravi Ramamoorthi and Ren Ng},
      year={2020},
      eprint={2003.08934},
      archivePrefix={arXiv},
      primaryClass={cs.CV},
      url={https://arxiv.org/abs/2003.08934}, 
}

@inproceedings{ViT,
   author = {Alexey Dosovitskiy and Lucas Beyer and Alexander Kolesnikov and Dirk Weissenborn and Xiaohua Zhai and Thomas Unterthiner and Mostafa Dehghani and Matthias Minderer and Georg Heigold and Sylvain Gelly and Jakob Uszkoreit and Neil Houlsby},
   booktitle = {ICLR 2021 - 9th International Conference on Learning Representations},
   title = {AN IMAGE IS WORTH 16X16 WORDS: TRANSFORMERS FOR IMAGE RECOGNITION AT SCALE},
   year = {2021}
}

@inproceedings{dpt,
   author = {René Ranftl and Alexey Bochkovskiy and Vladlen Koltun},
   doi = {10.1109/ICCV48922.2021.01196},
   issn = {15505499},
   booktitle = {Proceedings of the IEEE International Conference on Computer Vision},
   title = {Vision Transformers for Dense Prediction},
   year = {2021}
}

@InProceedings{ifrnet, 
  author = {Kong, Lingtong and Jiang, Boyuan and Luo, Donghao and Chu, Wenqing and Huang, Xiaoming and Tai, Ying and Wang, Chengjie and Yang, Jie}, 
  title = {IFRNet: Intermediate Feature Refine Network for Efficient Frame Interpolation}, 
  booktitle = {Proceedings of the IEEE/CVF Conference on Computer Vision and Pattern Recognition (CVPR)}, 
  year = {2022}
}

@inproceedings{eden,
  title={Enhanced Diffusion for High-quality Large-motion Video Frame Interpolation},
  author={Zhang, Zihao and Chen, Haoran and Zhao, Haoyu and Lu, Guansong and Fu, Yanwei and Xu, Hang and Wu, Zuxuan},
  booktitle={Proceedings of the IEEE/CVF Conference on Computer Vision and Pattern Recognition},
  year={2025}
}

@article{DiT,
  title={Scalable Diffusion Models with Transformers},
  author={William Peebles and Saining Xie},
  year={2022},
  journal={arXiv preprint arXiv:2212.09748},
}

@inproceedings{vgg,
   author = {Karen Simonyan and Andrew Zisserman},
   booktitle = {3rd International Conference on Learning Representations, ICLR 2015 - Conference Track Proceedings},
   title = {Very deep convolutional networks for large-scale image recognition},
   year = {2015}
}

@misc{sr,
      title={QuickSRNet: Plain Single-Image Super-Resolution Architecture for Faster Inference on Mobile Platforms}, 
      author={Guillaume Berger and Manik Dhingra and Antoine Mercier and Yashesh Savani and Sunny Panchal and Fatih Porikli},
      year={2023},
      eprint={2303.04336},
      archivePrefix={arXiv},
      primaryClass={eess.IV},
      url={https://arxiv.org/abs/2303.04336}, 
}

@inproceedings{colmap1,
    author={Sch\"{o}nberger, Johannes Lutz and Frahm, Jan-Michael},
    title={Structure-from-Motion Revisited},
    booktitle={Conference on Computer Vision and Pattern Recognition (CVPR)},
    year={2016},
}

@inproceedings{colmap2,
    author={Sch\"{o}nberger, Johannes Lutz and Zheng, Enliang and Pollefeys, Marc and Frahm, Jan-Michael},
    title={Pixelwise View Selection for Unstructured Multi-View Stereo},
    booktitle={European Conference on Computer Vision (ECCV)},
    year={2016},
}

@misc{posediff,
      title={PoseDiffusion: Solving Pose Estimation via Diffusion-aided Bundle Adjustment}, 
      author={Jianyuan Wang and Christian Rupprecht and David Novotny},
      year={2024},
      eprint={2306.15667},
      archivePrefix={arXiv},
      primaryClass={cs.CV},
      url={https://arxiv.org/abs/2306.15667}, 
}

@inproceedings{vggt,
  title={VGGT: Visual Geometry Grounded Transformer},
  author={Wang, Jianyuan and Chen, Minghao and Karaev, Nikita and Vedaldi, Andrea and Rupprecht, Christian and Novotny, David},
  booktitle={Proceedings of the IEEE/CVF Conference on Computer Vision and Pattern Recognition},
  year={2025}
}

@inproceedings{lpips,
   author = {Richard Zhang and Phillip Isola and Alexei A. Efros and Eli Shechtman and Oliver Wang},
   doi = {10.1109/CVPR.2018.00068},
   issn = {10636919},
   booktitle = {Proceedings of the IEEE Computer Society Conference on Computer Vision and Pattern Recognition},
   title = {The Unreasonable Effectiveness of Deep Features as a Perceptual Metric},
   year = {2018}
}

@inproceedings{TCL,
  title={Exploring motion ambiguity and alignment for high-quality video frame interpolation},
  author={Zhou, Kun and Li, Wenbo and Han, Xiaoguang and Lu, Jiangbo},
  booktitle={Proceedings of the IEEE/CVF Conference on Computer Vision and Pattern Recognition},
  pages={22169--22179},
  year={2023}
}

@inproceedings{Rangenullspace,
  title={Range-nullspace video frame interpolation with focalized motion estimation},
  author={Yu, Zhiyang and Zhang, Yu and Zou, Dongqing and Chen, Xijun and Ren, Jimmy S and Ren, Shunqing},
  booktitle={Proceedings of the IEEE/CVF Conference on Computer Vision and Pattern Recognition},
  pages={22159--22168},
  year={2023}
}

@inproceedings{Biformer,
  title={Biformer: Learning bilateral motion estimation via bilateral transformer for 4k video frame interpolation},
  author={Park, Junheum and Kim, Jintae and Kim, Chang-Su},
  booktitle={Proceedings of the IEEE/CVF Conference on Computer Vision and Pattern Recognition},
  pages={1568--1577},
  year={2023}
}

@inproceedings{rife,
  title={Real-Time Intermediate Flow Estimation for Video Frame Interpolation},
  author={Huang, Zhewei and Zhang, Tianyuan and Heng, Wen and Shi, Boxin and Zhou, Shuchang},
  booktitle={Proceedings of the European Conference on Computer Vision (ECCV)},
  year={2022}
}

@inproceedings{igs,
  title={Instant gaussian stream: Fast and generalizable streaming of dynamic scene reconstruction via gaussian splatting},
  author={Yan, Jinbo and Peng, Rui and Wang, Zhiyan and Tang, Luyang and Yang, Jiayu and Liang, Jie and Wu, Jiahao and Wang, Ronggang},
  booktitle={Proceedings of the Computer Vision and Pattern Recognition Conference},
  pages={16520--16531},
  year={2025}
}

@inproceedings{film,
  title={Film: Frame interpolation for large motion},
  author={Reda, Fitsum and Kontkanen, Janne and Tabellion, Eric and Sun, Deqing and Pantofaru, Caroline and Curless, Brian},
  booktitle={European Conference on Computer Vision},
  pages={250--266},
  year={2022},
  organization={Springer}
}

@inproceedings{kplanes,
  title={K-planes: Explicit radiance fields in space, time, and appearance},
  author={Fridovich-Keil, Sara and Meanti, Giacomo and Warburg, Frederik Rahb{\ae}k and Recht, Benjamin and Kanazawa, Angjoo},
  booktitle={Proceedings of the IEEE/CVF Conference on Computer Vision and Pattern Recognition},
  pages={12479--12488},
  year={2023}
}

@article{vimeo90k,
  title={Video Enhancement with Task-Oriented Flow},
  author={Xue, Tianfan and Chen, Baian and Wu, Jiajun and Wei, Donglai and Freeman, William T},
  journal={International Journal of Computer Vision (IJCV)},
  volume={127},
  number={8},
  pages={1106--1125},
  year={2019},
  publisher={Springer}
}

@article{ucf101,
  title={Ucf101: A dataset of 101 human actions classes from videos in the wild},
  author={Soomro, Khurram and Zamir, Amir Roshan and Shah, Mubarak},
  journal={arXiv preprint arXiv:1212.0402},
  year={2012}
}

@inproceedings{snu_film,
    author = {Choi, Myungsub and Kim, Heewon and Han, Bohyung and Xu, Ning and Lee, Kyoung Mu},
    title = {Channel Attention Is All You Need for Video Frame Interpolation},
    booktitle = {AAAI},
    year = {2020}
}

@article{nerfmm,
  title={Ne{RF}$--$: Neural Radiance Fields Without Known Camera Parameters},
  author={Zirui Wang and Shangzhe Wu and Weidi Xie and Min Chen and Victor Adrian Prisacariu},
  journal={arXiv preprint arXiv:2102.07064},
  year={2021}
}

@misc{instantsplat,
        title={InstantSplat: Sparse-view Gaussian Splatting in Seconds},
        author={Zhiwen Fan and Kairun Wen and Wenyan Cong and Kevin Wang and Jian Zhang and Xinghao Ding and Danfei Xu and Boris Ivanovic and Marco Pavone and Georgios Pavlakos and Zhangyang Wang and Yue Wang},
        year={2024},
        eprint={2403.20309},
        archivePrefix={arXiv},
        primaryClass={cs.CV}
      }

@article{noposplat,
      title   = {No Pose, No Problem: Surprisingly Simple 3D Gaussian Splats from Sparse Unposed Images},
      author  = {Ye, Botao and Liu, Sifei and Xu, Haofei and Xueting, Li and Pollefeys, Marc and Yang, Ming-Hsuan and Songyou, Peng},
      journal = {arXiv preprint arXiv:2410.24207},
      year    = {2024}
    }

@inproceedings{depthsplat,
      title   = {DepthSplat: Connecting Gaussian Splatting and Depth},
      author  = {Xu, Haofei and Peng, Songyou and Wang, Fangjinhua and Blum, Hermann and Barath, Daniel and Geiger, Andreas and Pollefeys, Marc},
      booktitle={CVPR},
      year={2025}
    }

@article{anysplat,
  title={Anysplat: Feed-forward 3d gaussian splatting from unconstrained views},
  author={Jiang, Lihan and Mao, Yucheng and Xu, Linning and Lu, Tao and Ren, Kerui and Jin, Yichen and Xu, Xudong and Yu, Mulin and Pang, Jiangmiao and Zhao, Feng and others},
  journal={ACM Transactions on Graphics (TOG)},
  volume={44},
  number={6},
  pages={1--16},
  year={2025},
  publisher={ACM New York, NY, USA}
}

@article{vru_data1,
  title={Swift4D: Adaptive divide-and-conquer Gaussian Splatting for compact and efficient reconstruction of dynamic scene},
  author={Wu, Jiahao and Peng, Rui and Wang, Zhiyan and Xiao, Lu and Tang, Luyang and Yan, Jinbo and Xiong, Kaiqiang and Wang, Ronggang},
  journal={arXiv preprint arXiv:2503.12307},
  year={2025}
}

@article{vru_data2,
  title={LocalDyGS: Multi-view Global Dynamic Scene Modeling via Adaptive Local Implicit Feature Decoupling},
  author={Wu, Jiahao and Peng, Rui and Jiao, Jianbo and Yang, Jiayu and Tang, Luyang and Xiong, Kaiqiang and Liang, Jie and Yan, Jinbo and Liu, Runling and Wang, Ronggang},
  journal={arXiv preprint arXiv:2507.02363},
  year={2025}
}
}

\clearpage
\setcounter{page}{1}
\maketitlesupplementary

\appendix

\section{Detailed Per-Scene Results}
\label{sec:App}
\begin{figure}[ht]
    \centering
    \includegraphics[width=\columnwidth]{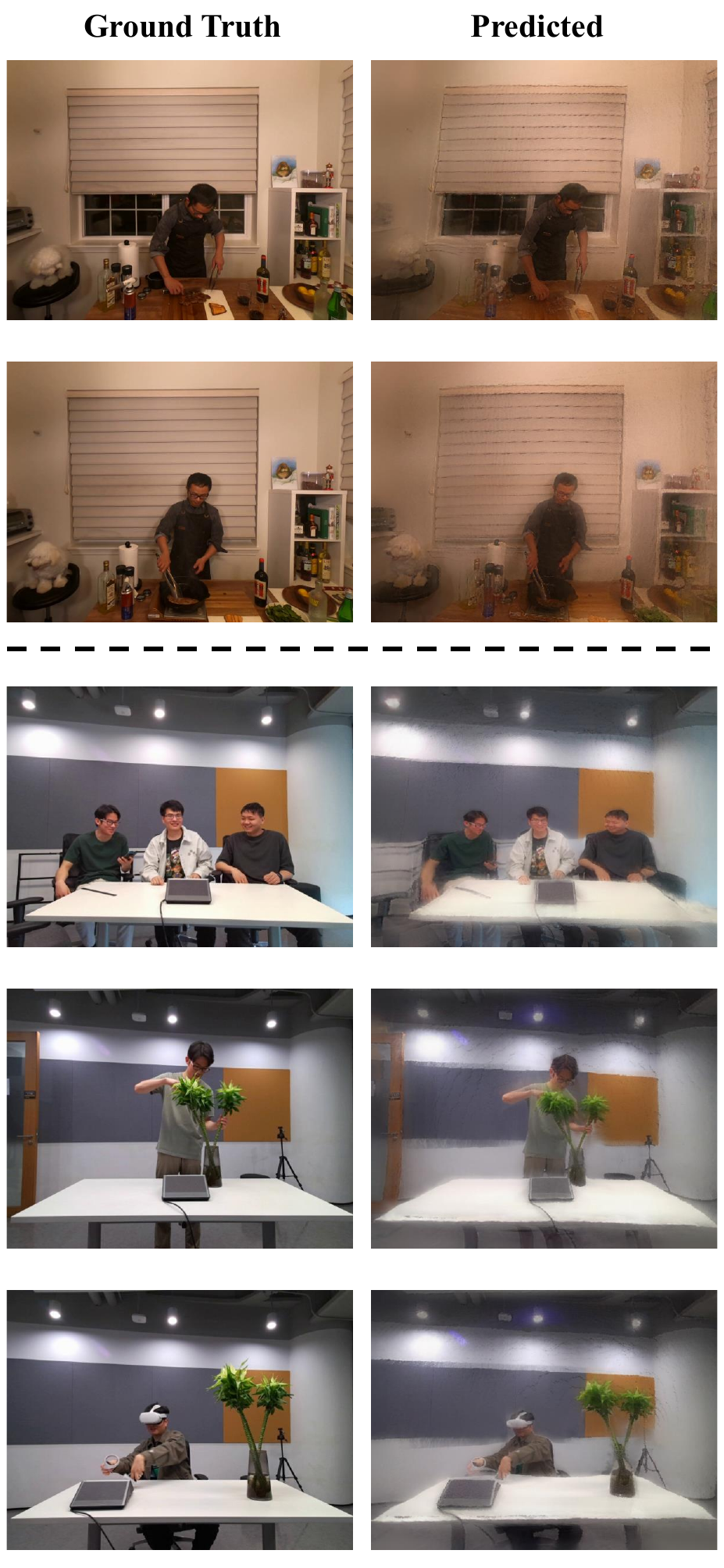}
    \caption{Qualitative comparison of reference and predicted viewpoints at frame 150 across all five scenes, rendered at $1024 \times 768$ resolution.}
    \label{fig:add_quali_res}
\end{figure}

In this section, additional qualitative and quantitative results are presented across all scenes rendered, using both Neural3DVideo~\cite{cooking_guy} and MeetRoom~\cite{meetroom} datasets. Leveraging 2 unposed input video streams to generate the output video from the target viewpoint at $1024 \times 768$ image resolution. \Cref{fig:add_quali_res} shows qualitative results for frame 150 of the generated video alongside the ground truth images. Per-scene quantitative results are summarized in \cref{tab:all_scenes}, where scores are averaged over all frames of each generated novel-view video.

\begin{table}[ht]
\centering
\caption{Quantitative results on the scenes described in \cref{subsec:datasets} using LiveStre4m, evaluated at two image resolutions in a single NVIDIA H100 GPU.}
\resizebox{\columnwidth}{!}{%
\begin{tabular}{l | l | c c}
\toprule
\textbf{Output Res. ($p$)} & \textbf{Scene} & \textbf{PSNR} $\uparrow$ & \textbf{Runtime (s)} $\downarrow$ \\
\midrule
\multirow{5}{*}{$1024 \times 768$} & cut         & 20.43 & 0.074 \\
                                   & sear        & 21.79 & 0.074 \\
                                   & discussion  & 18.75 & 0.074 \\
                                   & trimming    & 17.94 & 0.074 \\
                                   & vrheadset   & 19.24 & 0.074 \\
\midrule
\multirow{5}{*}{$512 \times 384$}  & cut         & 21.16 & 0.062 \\
                                   & sear        & 23.72 & 0.062 \\
                                   & discussion  & 20.92 & 0.062 \\
                                   & trimming    & 18.80 & 0.062 \\
                                   & vrheadset   & 18.24 & 0.061 \\
\bottomrule
\end{tabular}%
}
\label{tab:all_scenes}
\end{table}

%\begin{table}[h]
%\caption{Detailed performance of the LiveStre4m method in each individual scene in the test set.}
%\centering
%\resizebox{0.75\columnwidth}{!}{
%\begin{tabular}{l|cccc}
%\toprule
% Scene                                          & Runtime (s)$\downarrow$ & PSNR $\uparrow$  \\ 
%\midrule
% \textit{cut\_roasted\_beef}~\cite{cooking_guy} &    0.09                & 19.70 \\
% \textit{sear\_steak}~\cite{cooking_guy}        &    0.08                & 19.50        \\\midrule
% \textit{discussion}~\cite{meetroom}            &    0.08                & 11.50          \\
% \textit{trimming}~\cite{meetroom}              &    0.08                & 11.96         \\
% \textit{vrheadset}~\cite{meetroom}             &    0.08                & 11.57         \\
%\bottomrule
%\end{tabular}}
%\label{tab:detail_res}
%\end{table}

%\begin{figure}[b]
%    \centering
%    \includegraphics[width=\columnwidth]{sec/figures/appendix_split_sear.pdf}
%    \caption{Qualitative temporal results comparing expected and predicted viewpoints in the \textit{sear\_steak}~\cite{cooking_guy} scene.}
%    \label{fig:add_quali_res_sear}
%\end{figure}

\section{VRU-Basketball Dataset}
\label{sec:dynamic_data}
To validate the robustness of LiveStr4m in highly dynamic scenarios, it is evaluated on the VRU-Basketball dataset~\cite{vru_data1, vru_data2}. This dataset comprises multi-view recordings of professional basketball games, captured by 34 static cameras, providing a challenging benchmark due to rapid player motion and complex scene dynamics.

\begin{figure}[ht]
    \centering
    \includegraphics[width=\columnwidth]{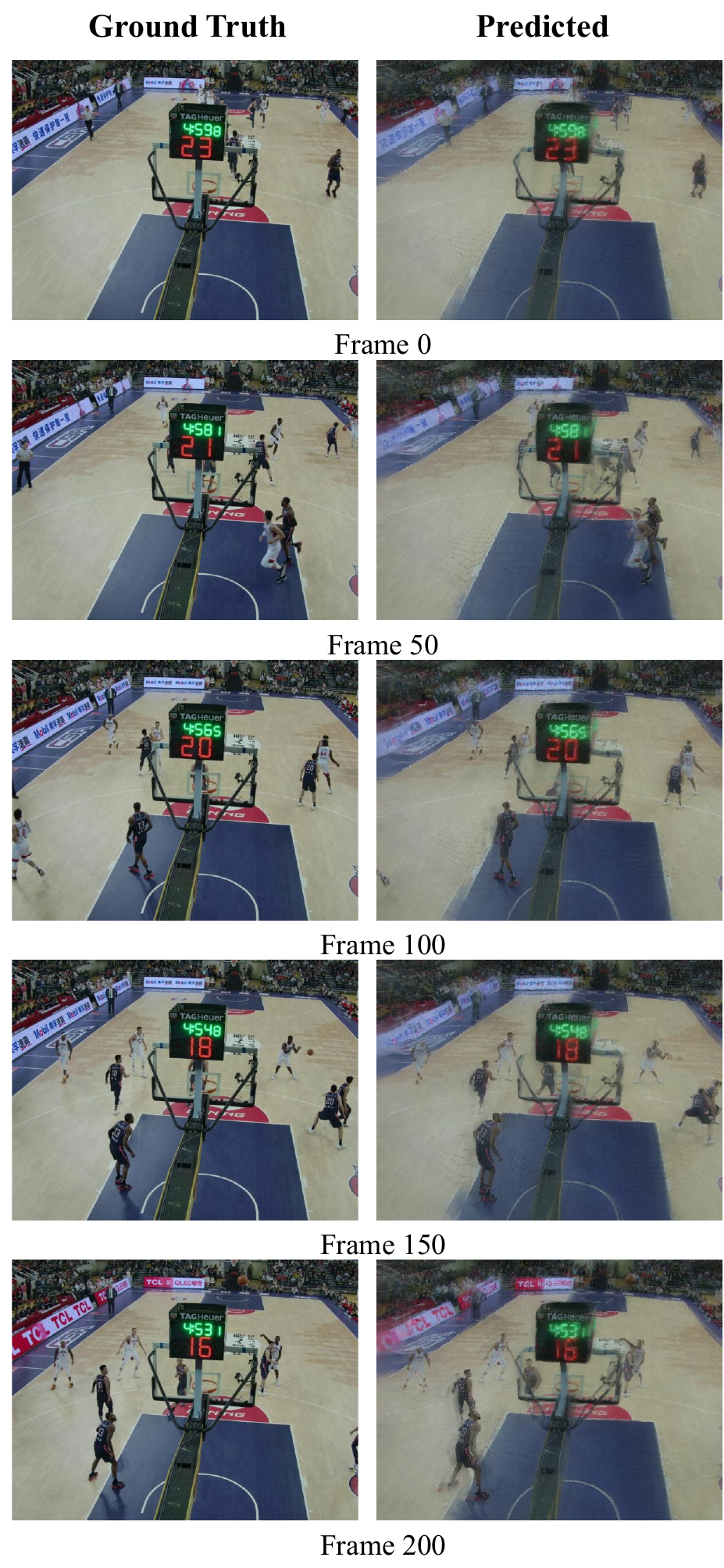}
    \caption{Qualitative results comparing different time steps of the generated video of the GZ scene of the VRU-Basketball dataset~\cite{vru_data1,vru_data2} with the expected outputs rendered at $1024 \times 768$ resolution.}
    \label{fig:vru}
\end{figure}

Consistent with the experimental setup described earlier in this paper, the central camera is selected as the target viewpoint, while  the two nearest cameras serve as input views. These unposed inputs are fed into LiveStre4m to generate the full video sequence from the target viewpoint. Quantitative results, including PSNR and runtime, are reported in \cref{tab:vru}, and qualitative comparisons between synthesized frames and ground-truth images are shown in \cref{fig:vru}.

\begin{table}[t]
\centering
\caption{Quantitative results on the VRU-Basketball dataset~\cite{vru_data1, vru_data2} obtained with two feed-forward methods, the proposed LiveStre4m (ours) and FLARE~\cite{FLARE}. Results obtained in a single H100 GPU.}
\resizebox{\columnwidth}{!}{%
\begin{tabular}{l | c c c}
\toprule
\multirow{2}{*}{\textbf{Model}} &
\multicolumn{3}{c}{\textbf{VRU-Basketball}} \\ 
\cmidrule(lr){2-4} 
 & Runtime (s) $\downarrow$ & PSNR $\uparrow$ & Resolution $\uparrow$ \\
\midrule
FLARE~\cite{FLARE}  & 0.248 & 17.42 & $512 \times 384$ \\
LiveStre4m (Ours)   & 0.076 & 17.88 & $1024 \times 768$ \\
LiveStre4m (Ours)   & 0.061 & 18.68 & $512 \times 384$ \\
\bottomrule
\end{tabular}%
}
\label{tab:vru}
\end{table}

\section{Pose Estimation Metrics}
\label{sec:pose}
Although LiveStr4m was not explicitly developed for camera pose prediction, this component plays an important role in downstream novel view synthesis. As described in in \cref{sec:method}, the estimated camera poses guide 3D scene reconstruction, making accurate predictions essential. \Cref{tab:pose} reports quantitative results for camera pose estimation, comparing LiveStre4m with FLARE~\cite{FLARE}. Employing standard camera pose accuracy metrics evaluated across the five scenes described in \cref{subsec:datasets}, LiveStr4m achieves performance comparable to the baseline, indicating that reliable pose estimation can be obtained even though it is not explicitly optimized for this task..

\begin{table}[ht]
\centering
\caption{Quantitative comparison of camera pose prediction accuracy. Metrics reported are RRA@$5^\circ$, RTA@$5^\circ$, and the combined AUC@$30^\circ$ (average of rotation and translation AUC).}
\resizebox{\columnwidth}{!}{%
\begin{tabular}{l | c | c c c}
\toprule
\textbf{Model} & \textbf{Resolution} & \textbf{RRA@$5^\circ$} $\uparrow$ & \textbf{RTA@$5^\circ$} $\uparrow$ & \textbf{AUC@$30^\circ$} $\uparrow$ \\
\midrule
FLARE~\cite{FLARE} & $512 \times 384$  & 100.00 & 60.00 & 91.66 \\
LiveStre4m         & $1024 \times 768$ & 100.00 & 60.00 & 92.09 \\
LiveStre4m         & $512 \times 384$  & 100.00 & 60.00 & 83.00 \\
\bottomrule
\end{tabular}%
}
\label{tab:pose}
\end{table}

\section{Real World Deployment}
\label{sec:deployment}

This paper shows that LiveStre4m is capable of generating high resolution novel-view videos at 13fps using a minimal buffer of only 3 input frames. However, several limitations remain for real-world deployment in live streaming scenarios, such as sports broadcasts or concerts. Namely, powerful hardware is required and the frame rate of the novel-viewpoint video is still lower than the desirable 24fps. Finally, as shown in \cref{tab:combined_results}, the visual quality of the generated video is lower than slower state-of-the-art 3D reconstruction methods. 

% 
%Having the supplementary compiled together with the main paper means that:
% 
%\begin{itemize}
%\item The supplementary can back-reference sections of the main paper, for example, we can refer to %\cref{sec:intro};
%\item The main paper can forward reference sub-sections within the supplementary explicitly (e.g. %referring to a particular experiment); 
%\item When submitted to arXiv, the supplementary will already included at the end of the paper.
%\end{itemize}
% 
%To split the supplementary pages from the main paper, you can use \href{https://support.apple.com/en-ca/guide/preview/prvw11793/mac#:~:text=Delete%20a%20page%20from%20a,or%20choose%20Edit%20%3E%20Delete).}{Preview (on macOS)}, \href{https://www.adobe.com/acrobat/how-to/delete-pages-from-pdf.html#:~:text=Choose%20%E2%80%9CTools%E2%80%9D%20%3E%20%E2%80%9COrganize,or%20pages%20from%20the%20file.}{Adobe Acrobat} (on all OSs), as well as \href{https://superuser.com/questions/517986/is-it-possible-to-delete-some-pages-of-a-pdf-document}{command line tools}.

\end{document}